\documentclass[a4paper, twocolumn]{article}

\usepackage{hyperref}
\usepackage{graphicx}  
\usepackage{fancyhdr}
\usepackage{amsmath}
\usepackage{amssymb}
\usepackage{enumitem}
\usepackage{subfigure}
\usepackage{tabularx}
\usepackage{colortbl}
\usepackage{booktabs}
\usepackage{array}
\usepackage{multirow}
\usepackage{url}
\usepackage{xcolor}
\usepackage{float}

\newcolumntype{D}{>{\centering\arraybackslash}X} 

\providecommand{\keywords}[1]
{
  \small	
  \textbf{\textit{Keywords---}} #1
}

\title{
  \bf Unlocking NACE Classification Embeddings with OpenAI for Enhanced Analysis and Processing. 
}

\author{
  A. Vidali, N. Jean, G. Le Pera\\
  \includegraphics[width = 40mm]{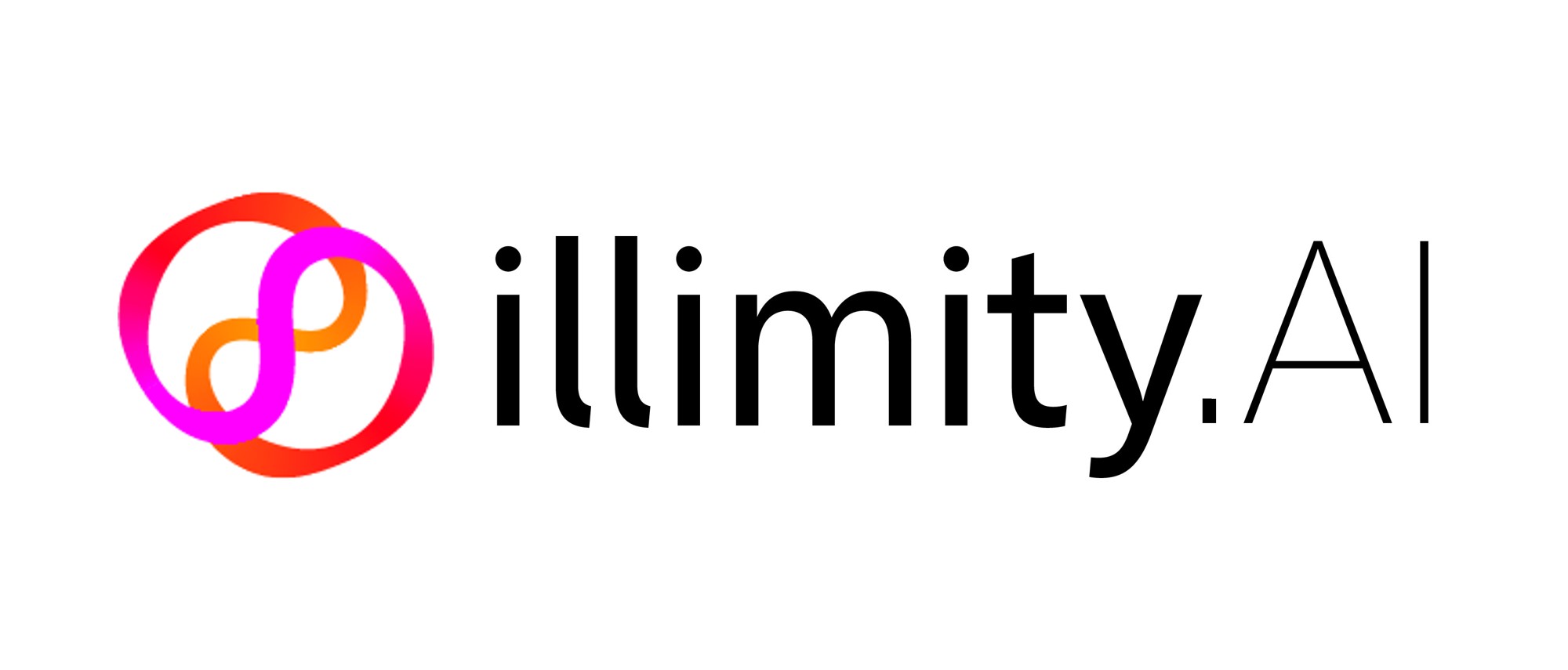}
}

\begin{document}

\maketitle

\begin{abstract}
  The Statistical Classification of Economic Activities in the European Community (NACE) is the standard classification system for the categorization of economic and industrial activities within the European Union. This paper proposes a novel approach to transform the NACE classification into low-dimensional embeddings, using state-of-the-art models and dimensionality reduction techniques. The primary challenge is the preservation of the hierarchical structure inherent within the original NACE classification while reducing the number of dimensions. To address this issue, we introduce custom metrics designed to quantify the retention of hierarchical relationships throughout the embedding and reduction processes. The evaluation of these metrics demonstrates the effectiveness of the proposed methodology in retaining the structural information essential for insightful analysis. This approach not only facilitates the visual exploration of economic activity relationships, but also increases the efficacy of downstream tasks, including clustering, classification, integration with other classifications, and others. Through experimental validation, the utility of our proposed framework in preserving hierarchical structures within the NACE classification is showcased, thereby providing a valuable tool for researchers and policymakers to understand and leverage any hierarchical data.
  \newline
\end{abstract}


\section{Introduction}\label{sec:intro}


The \textit{Statistical Classification of Economic Activities in the European Community}, commonly referred to as \textit{NACE}~\cite{statistical2008nace}, provides a standardized framework for categorizing European economic activities. Developed by the European Union, it organizes economic activities into a hierarchical structure based on the similarities of products, services, and processes. This hierarchical organization allows for the aggregation of related activities while facilitating sufficient differentiation of each category.

Despite its importance, the hierarchical nature of the NACE classification represents a significant obstacle to the effective use of standard analytical methods to process its data. Conventional approaches may be inadequate for capturing the complex relationships between economic activities and may overlook valuable information. In particular, traditional or naive encoding methods do not ensure that the relationships and semantic distances between NACE categories will be preserved in the encoded space.

In addition, the NACE classification has been updated over the years to reflect changes in the evolving economic landscape, but these updates often result in several categories becoming incompatible between different versions. Typically, regulators provide a mapping between old and new categories, but it often proves insufficient to map all categories between versions, and data continuity is compromised for many applications, especially in the context of machine learning.

Embedding models are crucial for translating the qualitative information preserved in classifications such as NACE into a numerical format that can be further processed. In this case, representing the NACE classification in the embedding form would have several advantages. For example, related categories will be positioned closer together, preserving their hierarchical and semantic relationships and allowing meaningful operations such as averaging or vector arithmetic to capture transitions between industries. The embedding representation also facilitates generalization beyond the predefined NACE categories and improves interpretability.

In summary, there is an increasing need for effective techniques to make the NACE classification usable in an analytical context while retaining its intrinsic structure and the relationships between its components. In context such as data analysis or machine learning, reducing the dimensionality of the embeddings is critical to further improve the usability of the NACE classification, especially for tree-based models such as decision trees or gradient boosting applied to tabular data. These models typically perform best with compact representations, as large and high-dimensional embeddings can lead to overfitting or increased computational complexity, ultimately compromising training convergence. Decreasing the size of the embeddings while preserving critical information ensures that these models can effectively process the data, leading to more accurate and interpretable results.

The primary objective of this paper is to propose a novel approach for transforming NACE classifications, or indeed any hierarchical classification, into high-dimensional embeddings and subsequently reducing their dimensions while preserving the hierarchical structure and, thus, the granularity of the underlying classification.

\section{Literature review}\label{sec:literature}

\subsection{NACE Classification}\label{subsec:nace}


The term NACE is derived from the French \textit{Nomenclature statistique des activités économiques dans la Communauté européenne}. It is a standardized framework developed by the European Union for categorizing economic activities at various levels of granularity. The principal objective is to establish a common language for classifying economic activities across different countries and regions, facilitating comparability and analysis.

The NACE classification is derived from the International Standard Industrial Classification (ISIC), which provides a universal standard for categorizing economic activities. NACE is a more detailed version of ISIC, with the same items at the highest levels, but with greater details at at the lower levels.

Since 1970, multiple NACE versions have been developed. The first version of the framework was developed in the 1970s, and the current revision, NACE Revision 2, was adopted in 2006. A newer minor version, denoted as NACE revision 2.1, was developed in 2023 and is set to be adopted from 2025 onwards~\cite{nacebackground}. This paper is based on the current NACE revision 2.

NACE revision 2 uses four hierarchical levels to identify each activity:

\begin{itemize}
  \item Level 1: 21 \textbf{sections} identified by alphabetical letters A to U.
  \item Level 2: 88 \textbf{divisions} identified by two-digit numerical codes (01 to 99).
  \item Level 3: 272 \textbf{groups} identified by three-digit numerical codes (01.1 to 99.0).
  \item Level 4: 615 \textbf{classes} identified by four-digit numerical codes (01.11 to 99.00).
\end{itemize}

This classification organizes economic activities into a multi-level hierarchy based on similarities in production processes, products, and services, enabling the grouping of related activities at higher levels while allowing for detailed differentiation within each category at lower levels. The full hierarchy of activities and their respective codes can be accessed via the official Eurostat dataset website by visiting \url{https://showvoc.op.europa.eu/#/datasets/ESTAT_Statistical_Classification_of_Economic_Activities_in_the_European_Community_Rev._2/data}

It should be noted that the first four digits of the code are the same in all European countries. However, national implementations may introduce additional levels.

\subsection{Embedding techniques}\label{subsec:embeddingtech}


Embedding techniques are widely used in machine learning and natural language processing to transform information, such as text, documents, images and audio, into a numerical representation.

Word embedding models, such as Word2Vec~\cite{mikolov2013efficient}, GloVe~\cite{pennington-etal-2014-glove}, and OpenAI's GPT (Generative Pre-trained Transformer) embedding models~\cite{openaiembeddings}, represent words as dense vectors in a continuous space based on their semantic meaning and context. These embeddings capture semantic relationships between words and have been successfully applied in document classification, sentiment analysis, and recommendation systems.

Also, graph embedding techniques aim to represent graph-structured data, such as social or citation networks, in a lower-dimensional space while preserving the structural information inherent to the data set. Methods like node2vec~\cite{grover2016node2vec} and GraphSAGE~\cite{hamilton2017inductive} learn embeddings for nodes in a graph by optimizing objectives that capture local and global network structures.

\subsection{Dimensionality reduction methods}\label{subsec:dimensionalityliter}


Dimensionality reduction techniques are essential tools for reducing the number of features or variables in high-dimensional datasets while preserving the essential information and structures.

PCA~\cite{wold1987principal} is one of the most widely utilized techniques for reducing the dimensionality of data sets. It transforms high-dimensional data into a lower-dimensional space while maximizing variance. By identifying the principal components that capture the most significant variations in the data, PCA enables the representation of complex datasets in a more compact form.

t-SNE~\cite{van2008visualizing} is a nonlinear dimensionality reduction technique commonly used for visualizing high-dimensional data in two or three dimensions. It works by minimizing the difference between the pairwise similarities in the high-dimensional and low-dimensional spaces. Unlike PCA, t-SNE preserves local similarities between data points, making it particularly useful for visualizing clusters and structures in complex datasets. However, t-SNE is known to be computationally expensive and sensitive to hyperparameters.

UMAP~\cite{mcinnes2018umap} is another nonlinear dimensionality reduction technique aiming to preserve both local and global structures in high-dimensional data. UMAP achieves this by constructing a low-dimensional embedding that represents the manifold structure of the data. Compared to t-SNE, UMAP is often faster and more scalable while producing similar or superior embedding quality.

While dimensionality reduction methods offer significant benefits in simplifying and visualizing complex datasets, they pose challenges, such as selecting appropriate techniques, tuning hyperparameters, and interpreting the resulting embeddings.

\subsection{Previous approaches}\label{subsec:previous}


Although the literature has not addressed the transformation of the NACE classification into embeddings, there are works associated with it that outline some of its issues.

As mentioned in section~\ref{subsec:nace}, NACE has undergone numerous iterations through various versions to remain aligned with the evolving social and economic landscape. This resulted in several challenges, including the complexity of aligning sections and categories across different versions. Perani et al.~\cite{perani2015matching} attempted to address this issue by constructing a conversion matrix for NACE levels. However, the fundamental challenge persists even in the context of future revisions. In this regard, this paper demonstrates that an approach whereby NACE descriptions are converted into embeddings has the potential to be scalable across versions.

With regard to the transformation of firm data into embeddings, Gerling~\cite{gerling2023company2vec} developed a system for translating the firm sector into embeddings using corporate websites as data sources. Gerling demonstrates the validity of this approach and its potential for a wide range of use cases in environments such as banks, including credit scoring, firm diversification, and peer identification, among others. This paper focuses on the subject of NACE rather than firm-based activity sectors, thus providing a more generally applicable approach.

\section{Methodology}\label{sec:methodology}

\subsection{NACE description preprocessing}\label{subsec:preprocessing}


The preprocessing of the NACE descriptions represents a fundamental step in the generation of high-quality embeddings. As previously stated, the NACE classification is hierarchical, meaning that each NACE description is a specialization of a more general NACE description, with the exception of the top level. Consequently, each NACE description carries two types of information: the description itself and the upper levels from which the description is derived.

For example, the NACE code 10.52 \textit{Manufacture of ice cream} is defined not only by its description but also by the descriptions of its upper levels:

\begin{enumerate}
  \item C \textit{MANUFACTURING}
  \item 10 \textit{Manufacture of food products}
  \item 10.5 \textit{Manufacture of dairy products}
  \item 10.52 \textit{Manufacture of ice cream}
\end{enumerate}

In order to obtain embeddings that retain the implicit hierarchical structure of each NACE element, each description has been concatenated with the descriptions of every upper level, if applicable. Additionally, all instances of the term "n.e.c." have been transformed to "not elsewhere classified".

Therefore, by applying such preprocessing, the description of code 10.52 \textit{Manufacture of ice cream} is transformed to: \emph{"Manufacturing. Manufacture of food products. Manufacture of dairy products. Manufacture of ice cream"}. This string serves as the input to the embedding model of NACE code 10.52.

In section~\ref{sec:results} the difference in metrics between this method and the raw NACE description without concatenation will also be demonstrated.

\subsection{NACE description embedding}\label{subsec:embeddingnace}


Several pre-trained models that transforms text into a numerical representation are available. This paper opted to test three different models: \textit{text-embedding-ada-002}~\cite{openaiada002}, \textit{text-embedding-3-small}~\cite{openai3smalllarge} and \textit{text-embedding-3-large}~\cite{openai3smalllarge} from OpenAI, which represents the latest models available at the time of development. The technical specifications of each models is presented in table~\ref{tab:modelslist}.

\begin{table}[htbp!]
  \setlength{\extrarowheight}{.75ex}
  \centering
  \begin{tabularx}{\linewidth}{XDDD}
    \toprule
    Model                  & Context length & Output dimension & Price per millon token (\$) \\
    \midrule
    text-embedding-ada-002 & 8191           & 1536             & $0.10$                      \\
    text-embedding-3-small & 8191           & max 1536         & $0.02$                      \\
    text-embedding-3-large & 8191           & max 3072         & $0.13$                      \\
    \bottomrule
  \end{tabularx}
  \caption{Specifications of embedding models}
  \label{tab:modelslist}
\end{table}

One of the main features of the newer models, Small and Large, is the capability to generate embeddings with shorter dimensions natively. This is made possible by a training technique~\cite{kusupati2022matryoshka} that OpenaAI employed for such models. With this functionality, which can be enabled by specifying the "dimensions" API parameter, the embeddings are shortened by removing elements from the end and normalizing the remaining portion. This feature will be evaluated in this paper by comparing it to other dimensionality reduction techniques described in section~\ref{subsec:dimensionalitymethod}.

The preprocessed descriptions of each NACE code were provided to the model through the official OpenAI API. As output, the model provided an embedding for each description of size 1536 for the Ada model and variable size for the Small and Large models.

\subsection{Dimensionality reduction}\label{subsec:dimensionalitymethod}


The default output size provided by the OpenAI embeddings model may be excessive for several applications. This is the rationale behind the utilization of dimensionality reduction techniques. This paper employs the dimensionality reduction algorithms of t-SNE and UMAP and also compares them to the native reduction offered by OpenAI models Large and Small. In general terms, the target dimension number for the reduction can be selected freely depending on the intended application, considering that fewer dimensions usually translate to less retained information from the original embedding. In this paper, the target dimension number for each reduction method is set to be 5. The selected hyperparameters for t-SNE are presented in Table~\ref{tab:tsne_hyperparameters}, and those for UMAP are presented in Table~\ref{tab:umap_hyperparameters}.

\begin{table}[htbp!]
  \centering
  \begin{tabularx}{\linewidth}{XD}
    \toprule
    Hyperparameter                & Value             \\
    \midrule
    dimensions                    & 5                 \\
    early exaggeration            & 5                 \\
    early exaggeration iterations & 1450              \\
    initial momentum              & 0.9               \\
    final momentum                & 0.9               \\
    iterations                    & 1480              \\
    perplexity                    & 11                \\
    metric                        & cosine similarity \\
    \bottomrule
  \end{tabularx}
  \caption{t-SNE hyperparameters}
  \label{tab:tsne_hyperparameters}
\end{table}

\begin{table}[htbp!]
  \centering
  \begin{tabularx}{\linewidth}{XD}
    \toprule
    Hyperparameter & Value             \\
    \midrule
    dimensions     & 5                 \\
    N of neighbors & 15                \\
    min dist       & 0.1               \\
    N of epochs    & 500               \\
    metric         & cosine similarity \\
    \bottomrule
  \end{tabularx}
  \caption{UMAP hyperparameters}
  \label{tab:umap_hyperparameters}
\end{table}

In our approach, we also explored the use of an Autoencoder~\cite{kramer1991nonlinear} to reduce the dimensionality of the embeddings generated by an embedding model. We aimed to preserve the essential features that capture the relationships between NACE categories in the high-dimensional embeddings by training the autoencoder to reconstruct the original embeddings from their compressed versions. The final results were suboptimal in comparison to the experiment presented in this paper, so we have decided not to include the results for the sake of clarity.

\subsection{Metric for structural retention}\label{subsec:metrics}


In order to assess the structural integrity of the NACE classification in the embedding form, it is essential to employ specific metrics. This paper employs two metrics to achieve this objective. The first is a metric commonly utilized in clustering, the \textit{silhouette}\cite{rousseeuw1987silhouettes}, to assess the degree of homogeneity of the embeddings. The second metric, entitled \emph{Hierarchy error}, has been created specifically for this study. It considers the interconnections between each related NACE code and measures how many remain valid in the embedding space. The following sections will provide a detailed description of these two metrics.

\subsubsection{Silhouette score}\label{subsubsec:silhouette}

The \textit{Silhouette score} is employed to measure the degree of homogeneity of the embeddings in the embeddings space, and its composed by calculating a \textit{Silhouette coefficient} for each NACE description's embedding. By definition, the Silhouette coefficient $s$ of a sample $i$ belonging to a clustered dataset is

\begin{equation}
  \text{s}(i) = \frac{b(i) - a(i)}{\text{max}(a(i), b(i))}
\end{equation}

where $a(i)$ is the mean distance between $i$ and all other samples in its own cluster, and $b(i)$ is the distance between $i$ and its next nearest cluster centroid. The outcome is bounded between $-1$ and $+1$, where $-1$ indicates incorrect clusters, and $+1$ indicates dense and well-separated clusters. Finally, the Silhouette score is produced by averaging all the Silhouette coefficients to produce a single metric on the whole dataset.

This paper employs the Silhouette score to measure the homogeneity of embeddings on three different clustering levels. In particular, for each level each embedding is labeled in the following manner, and then the Silhouette score is calculated for each level separately:

\begin{enumerate}
  \item \textbf{Level 1}: each code embedding is labelled according to its NACE \textit{section}.
  \item \textbf{Level 2}: each code embedding is labelled according to its NACE \textit{division}.
  \item \textbf{Level 3}: each code embedding is labelled according to its NACE \textit{group}.
\end{enumerate}

\begin{figure}[!htbp]
  \centering
  \includegraphics[width=\linewidth]{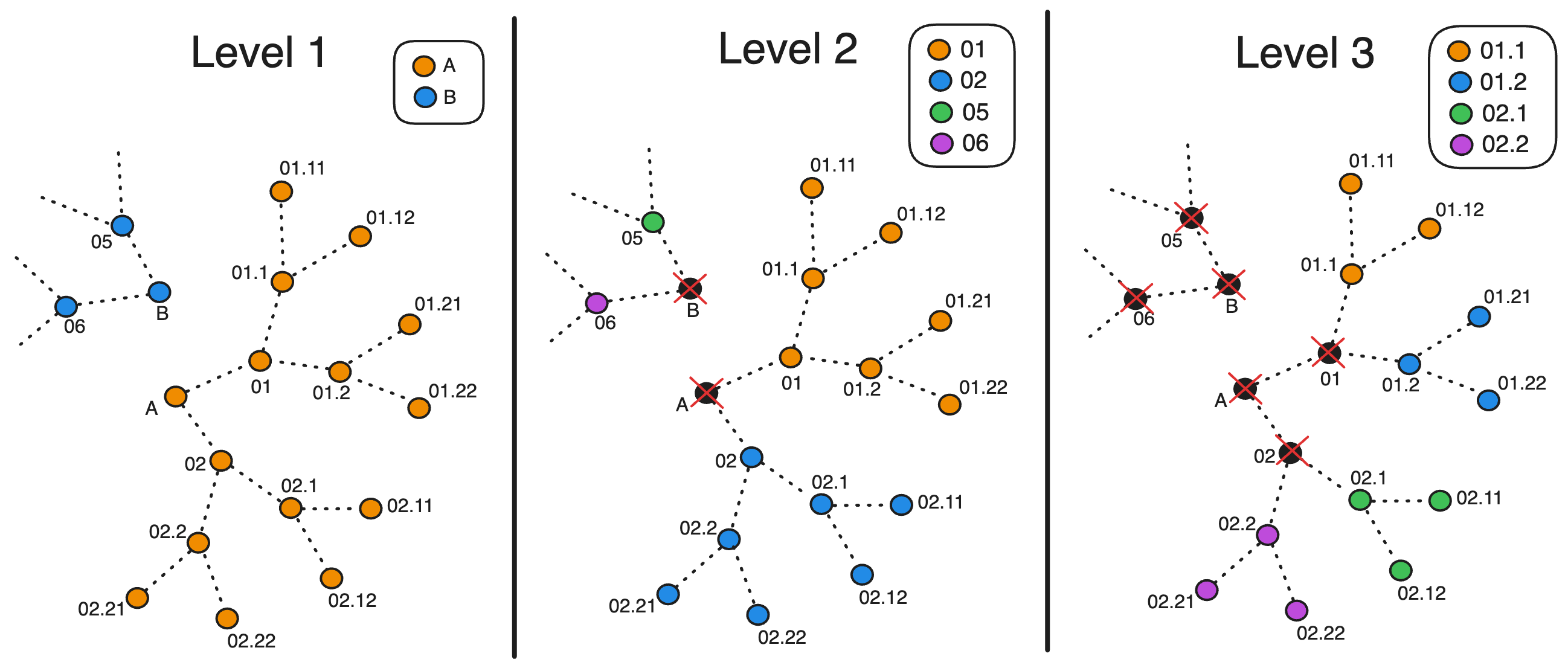}
  \caption{Silhouette labeling by level}
  \label{fig:silhouette_div}
\end{figure}

This method guarantees that each NACE level is taken into consideration when assessing the performance of the methods described. For levels 2 and 3, the silhouette calculation only includes the eligible code embeddings (i.e., code embeddings C \textit{MANUFACTURING} has no division or group within the hierarchical structure). Figure~\ref{fig:silhouette_div} illustrates a small example of such a clustering assignment by hierarchy level.

Ultimately, the three silhouette scores are averaged to yield a single score in order to aggregate the results at each level and facilitate performance comparison.

\subsubsection{Hierarchy error}\label{subsubsec:herror}

The Hierarchy error represents the percentage of NACE descriptions that have been misclassified. In this context, misclassified means that the embeddings of a code's description are not in the expected position in the embedding space.

For a clearer understanding of the metric, it's beneficial to view the NACE classification as a collection of trees. Each tree represents a specific NACE \textit{section} (A, B, C, and so on), with the section serving as the top node. Every \textit{division}, \textit{group}, and \textit{class} of that \textit{section} is connected consecutively, maintaining the original classification. In this context, the "children" of a NACE code are all the codes that originate from it, if they exist. Conversely, the "parent" of a NACE code is the broader category from which the code originates.

With the above setting, we introduce the following notation. Initially, the set of embeddings $E$ of each NACE code is defined as follows:

\begin{equation}
  E = embed(\text{NACE}_{d}) \ \forall \ d \in D
\end{equation}

where $D$ is the set containing all NACE descriptions, and $embed(\text{NACE}_{d})$ is the embeddings of the $d\text{-th}$ description.

Consequently, the Hierarchy error $\text{He}$ is defined as:

\begin{equation}
  \text{He} = \frac{\text{Hl}(e) >= 1 \ \forall \ e \in E }{|E|}
\end{equation}

where $\text{Hl}(e)$ represents the \textit{Hierarchy loss} of embedding $e$, a quantity that, if greater than 1, indicates that the embedding $e$ is misclassified. The Hierarchy error then represents the percentage of embeddings having a Hierarchy loss equal to or greater than 1.

The following section provides a detailed description of the Hierarchy loss.

\subsubsection{Hierarchy loss}\label{subsubsec:hloss}

The Hierarchy loss of a NACE embedding is a numerical representation of its position in the embedding space relative to its NACE upper level. If the loss is greater than 1, it indicates that the NACE code embedding has been misclassified with respect to the code's upper level. The Hierarchy loss is formally defined as follows:

\begin{equation}
  \text{Hl(e)} = \frac{\text{dist}(e, \text{parent}(e))}{\text{min}_{s \in \text{siblings}(\text{parent}(e)) }\text{dist}(e, s)}
\end{equation}

where:

\begin{itemize}
  \item $e$: the embedding of a NACE code.
  \item $\text{dist}(a, b)$: is the distance between embeddings $a$ and $b$, defined as $\frac{1 - \text{cosine}(a, b)}{2}$. If $a = b$, then $\text{dist}(a, b) = 0$.
  \item $\text{parent}(e)$: is the parent NACE code of the code having embedding $e$.
  \item $\text{siblings}(e)$: represent all NACE codes that share the same parent of the code having embedding $e$.
\end{itemize}

therefore:

\begin{itemize}
  \item $\text{dist}(e, \text{parent}(e))$: represent the distance between $e$ and the parent of the NACE code of $e$.
  \item $\text{min}_{s \in \text{siblings}(\text{parent}(e)) }\text{dist}(e, s)$: represent the smallest distance among all distances between $e$ and the siblings of the parent of the NACE code of $e$.
\end{itemize}

\begin{figure}[!htbp]
  \centering
  \includegraphics[width=\linewidth]{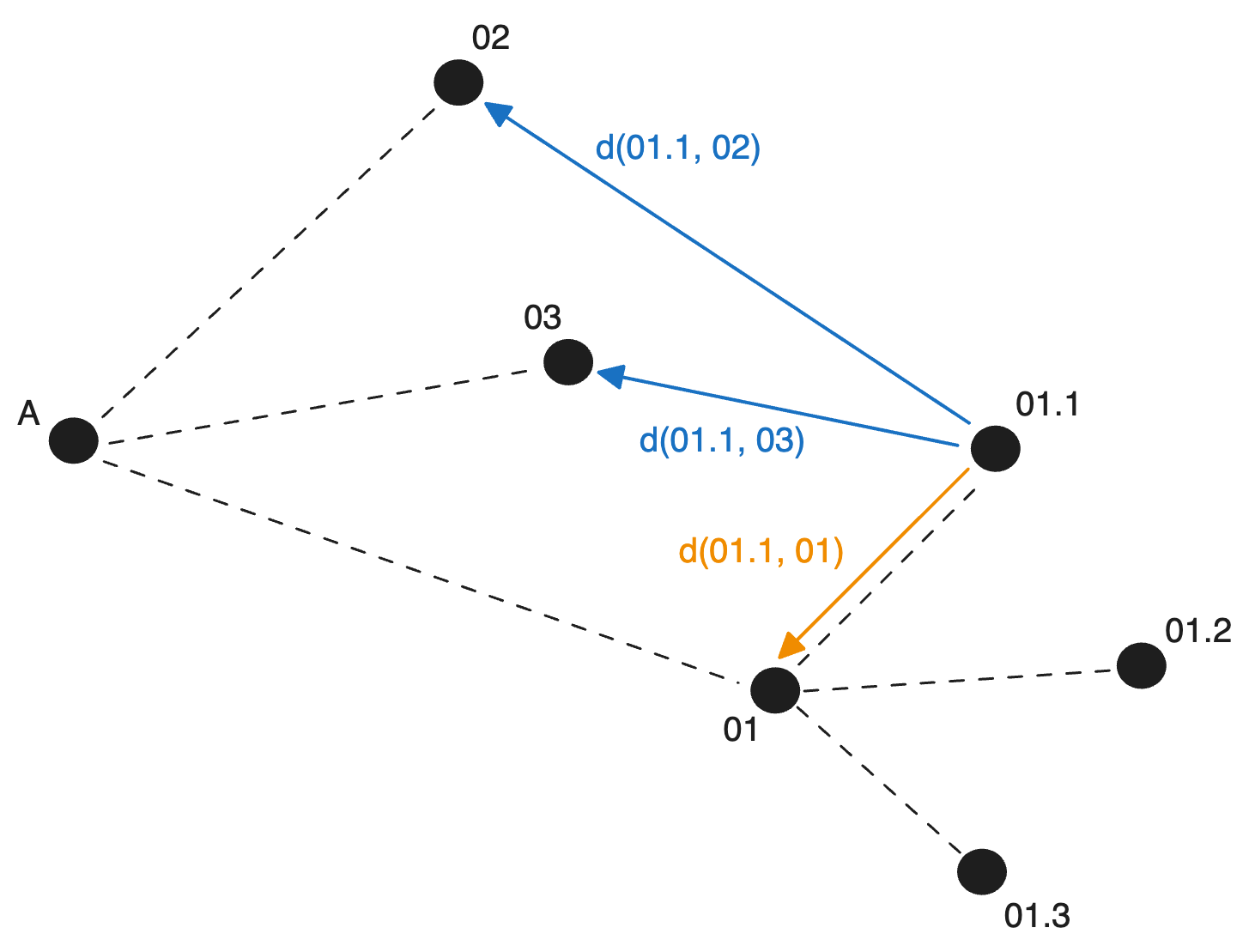}
  \caption{Hierarchy loss diagram.}
  \label{fig:hloss}
\end{figure}

Figure~\ref{fig:hloss} shows an example of the Hierarchy loss in a graphical representation.

The nodes represent a subset of the NACE codes. The dotted lines illustrate the hierarchical relationship between those codes. The orange arrow indicates the Hierarchy loss numerator of code 01.1, which represents the distance between code 01.1 and its parent code 01. The blue arrows indicate the distances between code 01.1 and its parent's siblings, namely codes 02 and 03. The Hierarchy loss considers the minimum of those distances, i.e., the shortest blue arrow, and compares it with the orange arrow. In the graphical representation, if the orange arrow is longer than the shortest blue arrow, the resulting Hierarchy loss will be greater than 1, meaning that 01.1 has been misclassified. In the ideal situation, a performant model would produce an embedding that places the node of code 01.1 closer to its parent than its parent's siblings in the embedding space, which would be the most desirable outcome.

It should be noted that this metric is not without its shortcomings. One limitation of the Hierarchy loss is that it does not affect the Hierarchy error unless greater than 1. This implies that the overall structure of the embedding nodes graph is not a primary concern as long as each code is closer to its parent than its parent's siblings. The other downside is the absence of a transitivity mechanism: if code 01.1 has a Hierarchy loss lower than 1 (relative to code 01), and code 01 has a Hierarchy loss lower than 1 as well (relative to code A), then it cannot be guaranteed that code 01.1 is also the closest of its level to A.

\section{Results}\label{sec:results}

\subsection{Full embeddings}\label{subsec:fullembeddings}


This section presents the results of the embeddings produced with and without the NACE description preprocessing, as introduced in section~\ref{subsec:preprocessing}. The embeddings utilized for this test are the full-size version, with those from the Ada and Small models having a size of 1536 and those from the Large model having a size of 3072. The nomenclature \textit{With parents} denotes the preprocessing with parent NACE codes description concatenated, whereas \textit{Raw} signifies that each NACE description has not undergone significant processing.

\begin{figure}[!htbp]
  \centering
  \includegraphics[width=\linewidth]{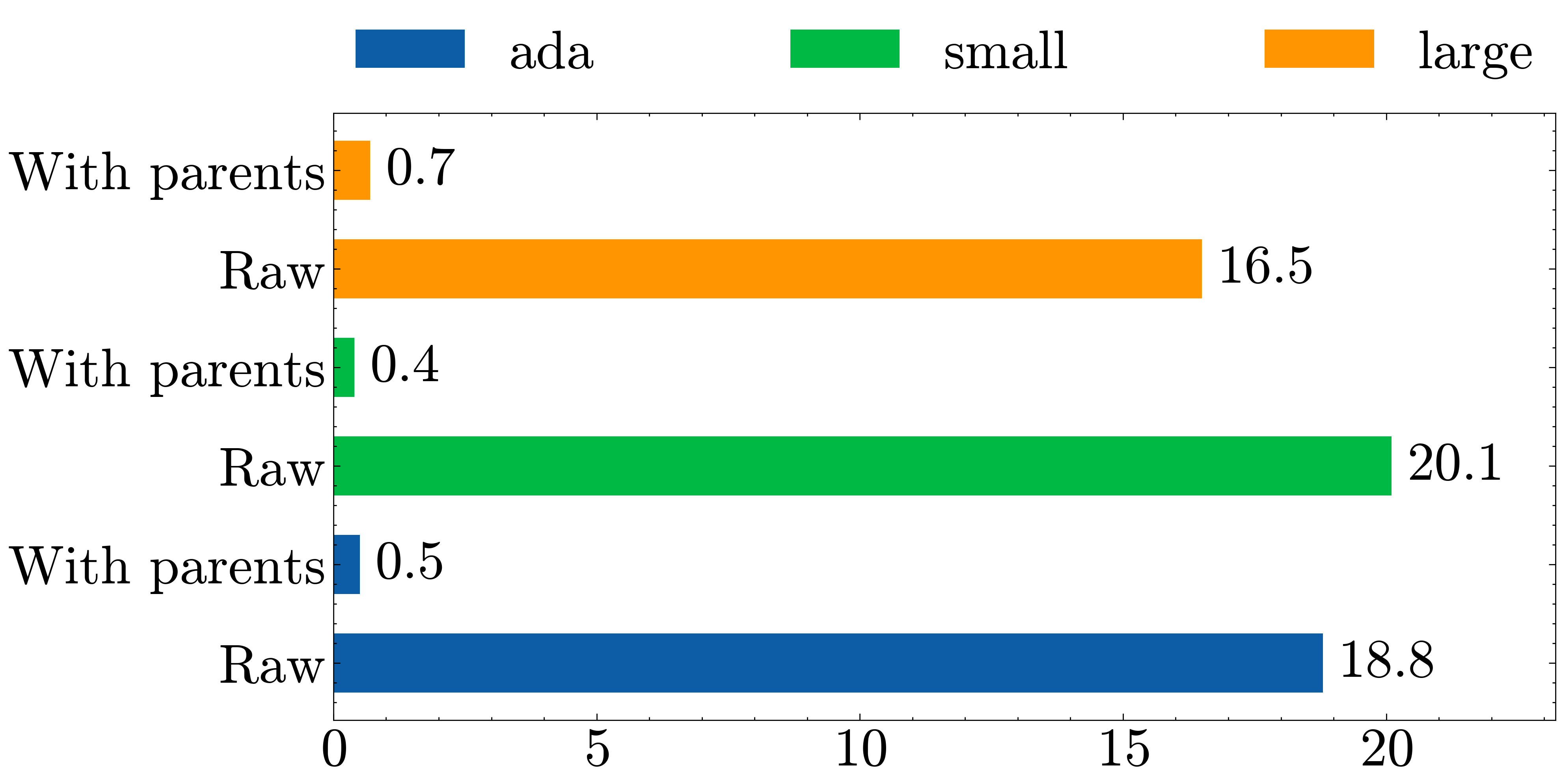}
  \caption{Hierarchy error on full embeddings. Lower is better.}
  \label{fig:herror_full}
\end{figure}

\begin{figure}[!htbp]
  \centering
  \includegraphics[width=\linewidth]{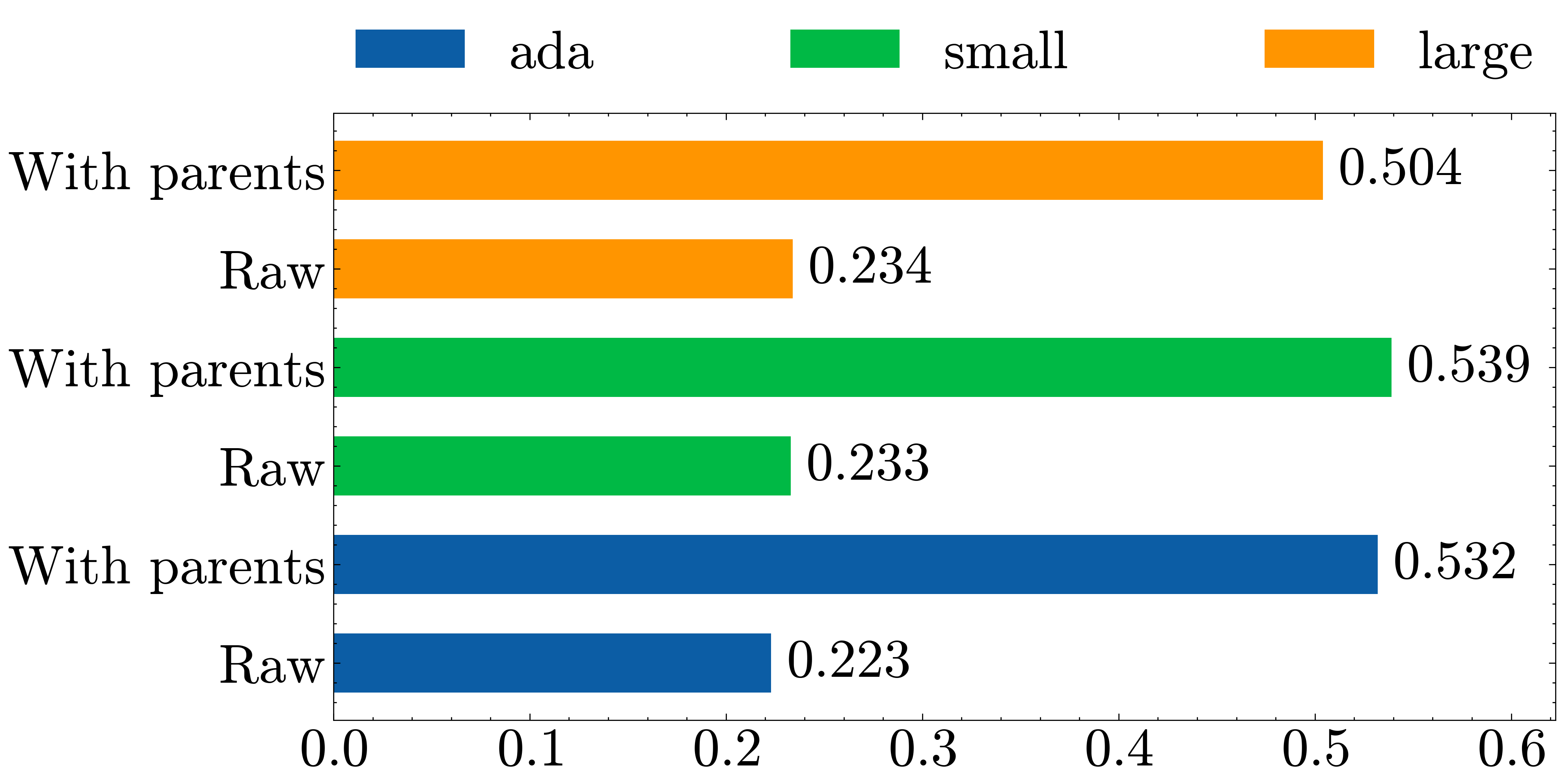}
  \caption{Mean silhouette on full embeddings. Higher is better.}
  \label{fig:silhouette_full}
\end{figure}

Figure~\ref{fig:herror_full} shows the Hierarchy error for the embeddings produced with full dimensions using the three OpenAI models, with and without the NACE description preprocessing. Figure~\ref{fig:silhouette_full} shows the Silhouette score on the same setup. It can be observed that the Hierarchy error is significantly lower when the information from parent codes is embedded in the code description. In particular, with the \textit{With parents} preprocessing, the Hierarchy error is lower than 1\%, while with the \textit{Raw} the misplaced NACE codes are almost 20\%. This is also evident in the Silhouette score graph, which shows more separated clusters for the enriched descriptions.

\subsection{Dimensionality reduction}\label{subsec:dimensionalityresults}


This section presents and analyzes the performance of the reduction techniques introduced in section~\ref{subsec:dimensionalitymethod}. As previously stated, each reduction technique modifies the embedding size from its original value to the chosen dimensionality of 5. For example, in the case of the Small model with t-SNE, the complete embeddings generated by the Small model of size 1536 are reduced to 5 dimensions through the application of the t-SNE algorithm.

The full embeddings used for the reduction process are exclusive of the \textit{With parents} type, as this preprocessing technique most effectively encodes the NACE classification structure.

In the figures, the nomenclature \textit{t-SNE} and \textit{UMAP} represent the reduction of a model's embeddings using such techniques. At the same time, \textit{Native} refers to the reduction produced natively through the OpenAI API by the Small and Large models described in section~\ref{subsec:embeddingnace}.

\begin{figure}[!htbp]
  \centering
  \includegraphics[width=\linewidth]{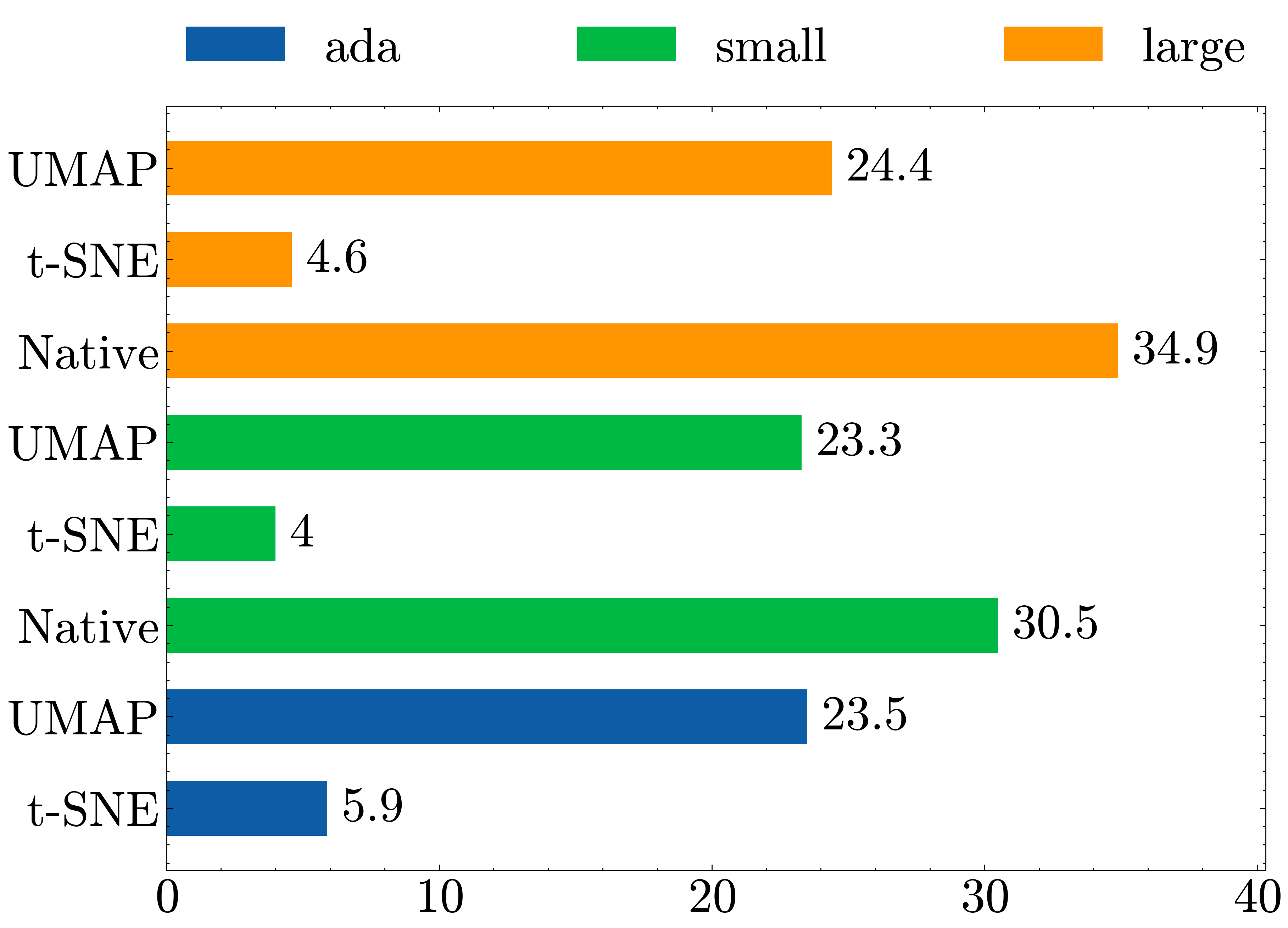}
  \caption{Hierarchy error on reduced embeddings. Lower is better.}
  \label{fig:herror_reduced}
\end{figure}

\begin{figure}[!htbp]
  \centering
  \includegraphics[width=\linewidth]{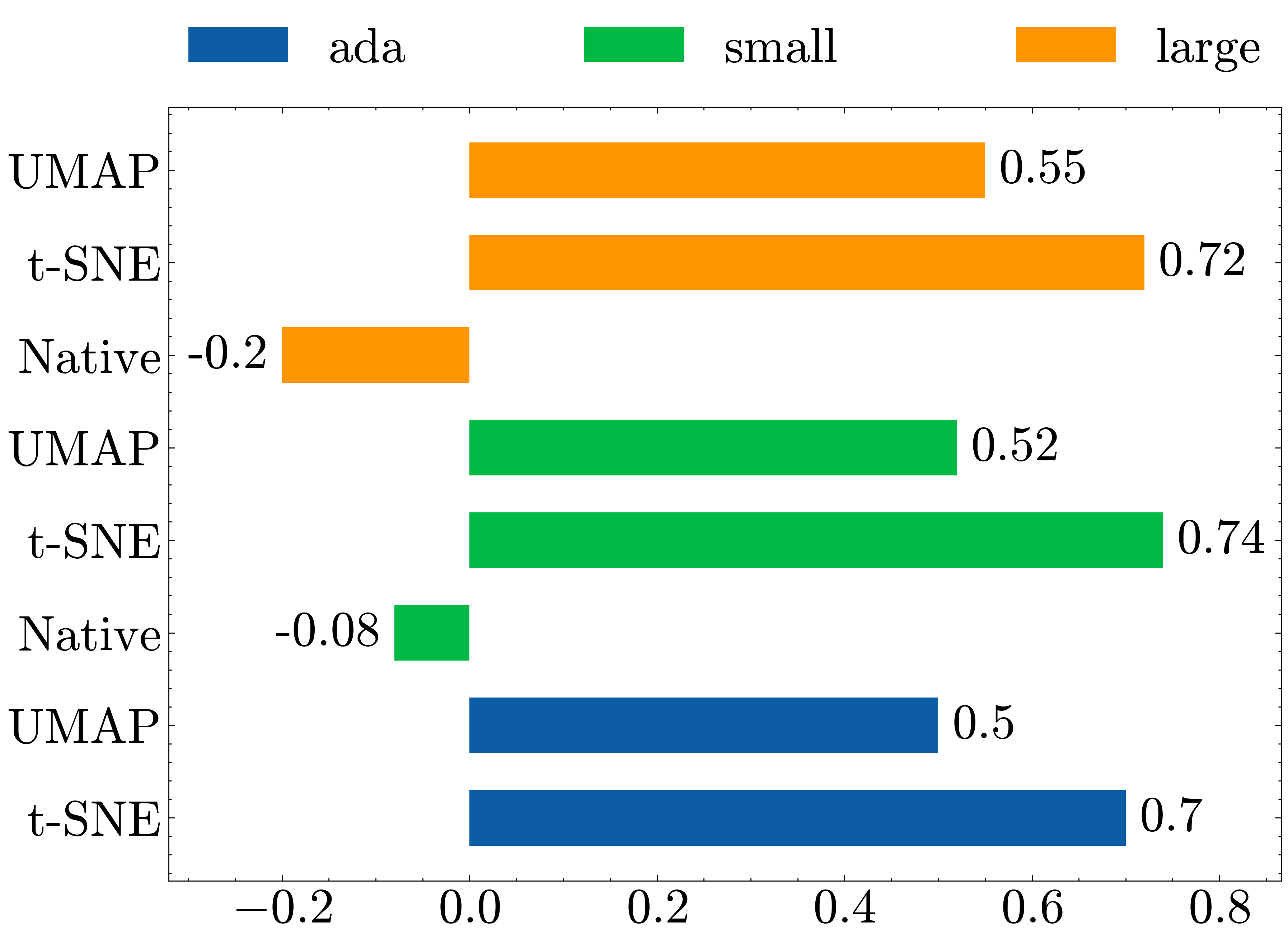}
  \caption{Mean silhouette on reduced embeddings. Higher is better.}
  \label{fig:silhouette_reduced}
\end{figure}

Figure~\ref{fig:herror_reduced} shows the Hierarchy error for the reduced embeddings of each model, while Figure~\ref{fig:silhouette_reduced} shows the Silhouette score for the same setup. The t-SNE algorithm yielded the better Hierarchy error, while UMAP exhibited worse performance. The native reduction by OpenAI models resulted in the most significant loss of information. Regarding the Silhouette score, t-SNE and UMAP demonstrated comparable results, with t-SNE exhibiting a more consistent edge. In contrast, the native reduction produces overlapping clusters. With respect to the performance of individual models, no model showed a clear advantage over the others in any metrics.

\subsection{Interpretation of results}\label{subsec:interpretation}


The findings presented in section~\ref{subsec:fullembeddings}, which assess the efficacy of the preprocessing phase, demonstrate that concatenating the descriptions of parents' NACE codes is a crucial element in encoding the original structure of the NACE classification in the embeddings. This conclusion depends significantly on the type of structure encoded; however, it is evident that a preprocessing of the input text is necessary to achieve optimal-quality embeddings.

Secondly, the results presented in section~\ref{subsec:dimensionalityresults} demonstrate the efficacy of reduction techniques, indicating that ad-hoc methods are more effective at retaining information compared to the native reduction from OpenAI. The native reduction method assumes that embedding values situated towards the end of the vector carry less information; consequently, by excluding these values and renormalizing the vector, the essential information is retained~\cite{openai3smalllarge}. In this case, the target size of 5 is likely insufficient for this technique to work optimally. In contrast, t-SNE and UMAP demonstrated robust performance, with t-SNE exhibiting a distinct advantage in the Hierarchy error metric. This is likely due to the algorithm's tendency to preserve local structure more effectively than global structure, which the Hierarchy error does not fully account for. Conversely, UMAP aims to balance global and local structures, which may result in a slight compromise in the quality of the local structure compared to t-SNE.

\subsection{Limitations and challenges}\label{subsec:limitations}


It's crucial to note that while this methodology has its strengths, it also comes with potential drawbacks and limitations.

First, it is challenging to identify a metric that effectively compares reduction methods while simultaneously accounting for the entirety of the local and global structure. While the Hierarchy error and the Silhouette score account for certain aspects, a more comprehensive evaluation would require the inclusion of additional metrics.

With this methodology, the functionality to incorporate additional samples into an existing set of embeddings is highly dependent on the reduction method employed. This emphasizes the need for careful consideration and planning, as the choice of method can significantly impact the process. For example, some implementations of t-SNE permit the incorporation of additional samples~\cite{Policar2024} while others don't~\cite{sklearntsne}. In the context of the NACE classification, this issue is not a significant concern given the long revision cycle, which spans years or even decades. However, in other applications, it could have a more immediate impact.

Finally, the rapid development of embedding models and LLMs in recent years, not only by OpenAI but also by other key players, is a significant factor in the longevity of the current models. In the coming months or years the performance of current models could be significantly surpassed, necessitating the generation of novel embeddings to enhance the overall methodology's performance.

\section{Conclusions}\label{sec:conclusions}


This paper proposes a methodology for transforming a qualitative classification, such as the NACE classification, into a numerical representation while retaining the original structure, along with methods to evaluate the efficacy of such a transformation. The first key aspect proposed is the preprocessing of the input text in order to incorporate the information regarding parent elements into the text of each element. The second key aspect is the employment of an embedding model with state-of-the-art performance, such as the \textit{text-embedding-3-small} and \textit{text-embedding-3-large} models from OpenAI. Subsequently, a dimensionality reduction method is essential if the downstream task requires embeddings with a smaller dimensionality. In such instances, t-SNE and UMAP effectively reduce the embedding dimensionality while maintaining most of the embedding information. Additionally, two metrics are proposed to assess the embedding performance throughout the pipeline.

This methodology has several applications that can benefit from using quality embeddings. Regarding the NACE classification, these include the incorporation of features into machine learning models, similarity search, clustering and economic sector analysis, the mapping of NACE classification to others like \textit{NAICS}~\cite{naics2024} and \textit{ISIC}~\cite{isic2008}, and the development of recommendation systems.

Future research will focus on further enhancing the results achieved so far while also, in the longer term, seeking to capitalize on the NACE classification embeddings generated with this framework. Regarding the results, potential approaches for further investigation include the incorporation of more significant data into the embedding model's input text to generate more comprehensive and informative representations. Additionally, the introduction of a novel metric that more accurately assesses the embeddings quality, both in terms of local and global evaluation with respect to the original classification structure, could prove beneficial. With regard to further analysis of the embedded NACE classification, it would be valuable to employ such data to improve the understanding of the current global economic dynamics. For instance, the use of embeddings could facilitate cross-country and cross-industry comparisons or assist in the identification of early signals of emerging economic sectors.

\section*{Acknowledgments}

We would like to extend our gratitude to Claudio Nordio for his role as Chief Risk Officer of illimity Bank, which supported the overall progress of this work.

\bibliographystyle{plain}
\bibliography{biblio}

\end{document}